# Assistive System in Conversational Agent for Health Coaching: The CoachAI Approach

Ahmed Fadhil

University of Trento, Trento, Italy
ahmed.fadhil@unitn.it

Abstract. With increasing physicians' workload and patients' needs for care, there is a need for technology that facilitates physicians work and performs continues follow-up with patients. Existing approaches focus merely on improving patient's condition, and none have considered managing physician's workload. This paper presents an initial evaluation of a conversational agent assisted coaching platform intended to manage physicians' fatigue and provide continuous follow-up to patients. We highlight the approach adapted to build the chatbot dialogue and the coaching platform. We will particularly discuss the activity recommender algorithms used to suggest insights about patients' condition and activities based on previously collected data. The paper makes three contributions: (1) present the conversational agent as an assistive virtual coach, (2) decrease physicians workload and continuous follow up with patients, all by handling some repetitive physician tasks and performing initial follow up with the patient, (3) present the activity recommender that tracks previous activities and patient information and provides useful insights about possible activity and patient match to the coach. Future work focuses on integrating the recommender model with the CoachAI platform and test the prototype with patient's in collaboration with an ambulatory clinic.

Keywords: Health coaching, conversational agents, recommender systems

## 1 Introduction

The prevalence of chronic diseases is due to the heterogeneity in individuals' demographics and their lifestyle pattern. Chronic conditions are associated with individuals living condition and behaviour as factors contributing to the escalation of this burden. Efficient and cost-effective follow-up is needed to decrease physician's involvement and guarantee continues patient's care. Tracking patients' condition through health coaching telemedicine systems is effective to combat the above burden [4]. The provision of eHealth and telemedicine services to stimulate patient's self-efficacy is a promising paradigm to reduce the burden of patient's care on the healthcare system [6]. Despite numerous technology mediated attempts to promote patient's lifestyle, existing interventions focus on tracking patient's condition and hoping they follow physician recommendations to improve their condition. Studies are focused on technical limitations of existing systems and few have considered decreasing physicians workload caused by patient's need for care. There is no way yet to fully connect physicians and patients, such that physicians access timely information about patients' condition and hints about possible actions. In this paper, we present CoachAI, an e-coaching platform for health and wellness, assisted by a chatbot. The recommender system in CoachAI is the main point we tend to investigate in this study. We discuss the design case, focusing on a holistic development of a personalized chatbot coach that handles some physician tasks and provides self-management training. The platform is formed by three separate parts, namely a web application, a chatbot dialogue system and an activity evaluator and recommender. The platform focuses on activity recommendation, decrease physician workload, and recommending actions to be taken towards promoting patient's health. Finally, rather than recommending activities to patients, we focus on analyzing patient condition, their previous activities and performance history, then conclude recommendations about physical activity and diet and provide them to the physician. We believe this will help decrease the workload for the physicians and provide precise patient's condition monitoring.

## 2  Background

Promoting health and wellness is a holistic approach to maintain the overall wellbeing of the nation. Three types of coaching techniques exists in telemedicine applications, fully manual (e.g., commands are provided and tracked by a human, which presents no intelligence at all), fully automated (e.g., commands are predefined and automatically assigned and tracked, this type is complicated to build) and semiautomated (e.g., commands are provided and tracked by either a human or a system or a negotiated between the two, this system is a negotiation between the previous two and substitutes the task between the software and a human). A previous work by Fadhil et al., [2] highlighted the importance of human actor in the loop while tracking patients' condition. Despite huge number of eHealth apps in the market, there is little research focused on user experience and views on a wide range of features. Health promotion is the main objective of primary care interventions. However, monitoring whether a patient follows a plan and if the plan fits with their condition is hard to track by physicians. There exist approaches to recommend caloric amount to consume per day, distance to run, amount of sleep to get. These are often recommended to the user to follow. However, user motivation to perform these activities is important to achieve long-term engagement. Supporting eHealth coaching with a real human makes a difference [2]. Effective coaching and tailored feedback in terms of its timing, content and interaction design are crucial elements in affecting behavioral change. We believe providing insights about patient's condition and performance to the physicians is helpful to reshape and personalize the plan to fit with patient's condition. The new generation of lifestyle coaching systems should predict optimal timing for feedback based on previously given performances. A personal plan has to be challenging and reachable, step by step, within timeframe. A work by Lim et al., [5] discussed the development of a wellness recommender system that helps individuals adapt suitable personalized wellness therapy treatments. The approach recommends wellness activities to users based on solutions from similar previous cases. This simple model provides personalized wellness recommendations. Another work by Fadhil et al., [3] discussed integrating a chatbot application to provide behaviour change interventions in telemedicine for healthy lifestyle promotion. The bot provides health support to nutrition education that helps overcome current limitations of similar systems. Conversational agent in the era of telemedicine systems is a novel approach to engage users in a task through a conversation. Although its new emergence, chatbots existed for many years, however they were limited in terms of capabilities. With the rise of deep learning and neural networks, it became possible to build complex domain dependent dialogue systems. A work by Shang et al., [7] proposed a Neural Responding Machine (NRM), a neural network-based response generator for Short-text Conversation. The NRM is trained with a large amount of one-round conversation data collected from a microblogging service. The findings revealed the system could generate grammatically correct responses over 75% of the input text.

## 3  CoachAI Platform

Being a conversational assisted e-coaching platform, CoachAI engages users into health-related activities to promote their lifestyle through a conversation medium. This application lets users (patients) to select their preferences by chatting with a bot, before assigning an activity, and accessing it. The conversational agent is a medium of interaction for patient's and the web application is for healthcare providers to track patient's performance. The activity recommender is to assist physician with patient clustering and activity assignment. This is achieved by providing physicians insights about patient's condition in relation to previous activities and similar patients. CoachAI is domain independent and general in terms of architecture to integrate into any domain falling into the context of health and wellbeing. The system tracks patient's performance report it to the coach, and most importantly assist the coach with activity assignment by analyzing aspects of previous patient interaction and building a recommender system to inform the coach about patients' group and condition.

### 3.1  Coach Portal

This is a web dashboard for the physician to initially register a patient and create their profile. The coach will be provided with further data about patient's adherence and compliance with a given activity. Moreover, the activity recommender will help the coach by clustering new patient's into groups, assist them with activity assignments, based on previous patient's and activity history.

### 3.2  Chatbot Agent

Once chatbot accessed, the patient begins a conversation about specific topics defined in a structured way through the dialogue. This step gathers data about user's BMI, physical activity, diet, sleep, and profession. The concept of conversational agents has already been applied to improve aspects of user lifestyle, through face-to-face conversation, with an actual agent using speech and hand gesture [1]. The conversational agent in CoachAI is based on text conversation and simple graphical elements embedded into the conversation. The bot performs two separate phases, namely information gathering and feedback collection. Initially, the bot chats with patient's and asks questions about their dietary pattern, physical activity, and other personal health related information. The bot then begins by asking user's feedback about a given activity. The bot performs a follow-up question to investigate aspects of patient's motivation and mental state during activity performance. All patient feedback is submitted to the coach, to notify them about patient condition. The feedback is also fed into an activity recommender to tweak existing activity data and patient performance. This is to increase recommendation accuracy prediction.

Chatbot Platform To provide patient's with an interface to communicate with the bot, we use Telegram cloud-based instant messaging service API[1]. The API provides security, and advanced CUI functionalities. The chatbot handles tasks about health information, booking an appointment, and tracking patient condition and various lifestyle related activities (see Figure1). Chatbots offer simplicity, adaptiveness, constant presence and motivation to patient's while interacting.

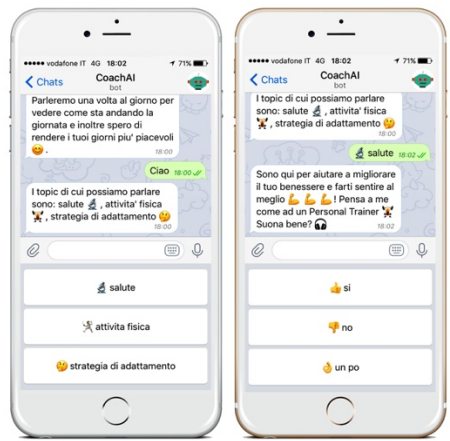

Fig. 1: The CoachAI Telegram Interface.

Conversation Dialogue The dialogue is based on a structured decision tree that is built using ChatScript[2], a chatbot engine that offers advanced features and capabilities that permit extremely clever chatbot behaviour programming. The ChatScript also offers useful ontology of nouns, verbs, adjectives, and adverbs for meaning extraction, see example below.

```
Topic: ~food(    ~fruit    fruit food eat)
t: What is your favorite food?
```

---

[1] https://telegram.org/
[2] http://chatscript.sourceforge.net/

```
        a: (~ fruit) I like fruit also.
        a: (~ metal) I prefer listening to heavy metal music rather than eating
            it.
?:    WHATMUSIC ( << what music you ~like >>) I prefer rock music.
s:  ( I * ~like * _~ music_types )
```

Based on the defined pattern, ChatScript allows constructing conversation pattern about various topics involved in the conversation. Below is an example of a conversation between the CoachAI agent and a patient.

```
...
User: Healthy Diet
CoachAI: I am here to help with your wellbeing and get you feeling your very
    best
Think of me as your virtual physician , sound good?
User: yes
...
```

### 3.3 Activity Recommender

This is formed by diet and physical activity recommendation. Initially, the bot collects data about patients' dietary pattern, BMI and user diet related data. This configures the impact factor for recommending personalized diet. The dietary recommendation is provided to the coach to help assist them with assigning a dietary activity. The physical activity recommender relies on an ontology to detect activity types and closeness among patient's from performance perspective. Physical activity types that are closer in their intensity are considered in the recommender model. Based on the activity metrics and closeness among patients, we predict the activity that fits a new patient's condition.

Initial Patient Clustering The system initially clusters patient's into groups with others that share similar lifestyle, health condition, activity performance, profession, ...etc. Whenever the coach registers a patient, they provide their personal data to the system through the conversational chatbot dialogue. This data consists of patient's health record, BMI, profession, dietary pattern and physical activity level. All this information is fed into the activity recommender model to help cluster patient's with similar features. This assists the coach to understand a patient's position with respect to patient activity hierarchy. The algorithm to perform patient clustering is highlighted in the Algorithm1 code snippet below. The algorithm performs iterative refinement approach following the steps listed:

1. The algorithm defines initial patient group centroids, which is achieved in different strategies. A common approach is assigning random values for centroids of all patient groups. However, since lifestyle information provides a strong predictor about patient's condition, before checking other features for the first two clustering approaches, we rely on patient's adherence to assign centroid of patient groups (high adherence and different adherence).

## Algorithm 1: Initial Patient Clustering

```
1  PatientClustering (P; C)
     inputs  :P={p_1; p_2; ...; p_n}(set of patient's to be clustered)
2            K(number of clusters);
3            MaxIters(Limit of iteration);
     output  :C={c_1; c_2; ...; c_k}(set of cluster centroids)
4    { L = {l(p) | p = 1; 2; ...; n} (set of cluster labels of P);
5    };
6    foreach c_i ∈ C do
7    |   c_i ← p_i ∈ P;
8    end
9    foreach p_i ∈ P do
10   |   l(p_i) ← argminDistance(p_i; c_j) j ∈ {1; ...; k};
11   end
12   Changed ← false;
13   Iter ← 0;
14   repeat;
15   foreach c_i ∈ C do
16   |   UpdateCluster (c_i);
17   end
18   foreach p_i ∈ P do
19   |   minDist ← argminDistance(p_i; c_j) j ∈ 1; ...; k;
20   |   if minDist ≠ l(p_i) then
21   |   |   l(p_i) ← minDist;
22   |   |   Changed ← true;
23   |   end
24   end
25   iter + +;
26   Until Changed = true && iter ≤ MaxIters;
```

2. This step assigns each patient to a cluster with the closest centroid. To find the cluster with similar centroid, the algorithm should calculate the distance between all patient's and each centroid.
3. This step recalculates the values of the centroids. The value's fields are updated, by averaging the values of patient's attributes that are part of the cluster.
4. Repeat steps 2 and 3 iteratively until patient's groups are fixed and stable.

Physical Activity Recommendation This is calculated from the activity matrix and user closeness in the feature space. Based on these measures, we cluster users into groups of active, moderate and passive users. When a new user is registered, the model calculates their information and feeds it into the model. Based on their closeness in the feature space, the model decides the group of users where it fits. The model then recommends an activity that could fit that user (patient). The activity recommender is to help the coach with decision making from a complex set of information collected from patients to recommend a physical activity. We define the similarity among patients, based on general recommendations for healthy lifestyle data. Based on expert support, and detailed information of different activities we build ontology trees and tables for different attributes of physical activities. These ontologies are used to calculate distances of different physical activities. The recommender system is based on actual physical activity data of people with positive exercise history. This recommends to people with similar features and who have sedentary lifestyle. The lifestyle information collected about a patient, might be fairly good indicator to show how similar people interact. Hence, when a new patient joins, the recommender finds a similar person, with an exercise history, and based on coach's decision, recommend a similar exercise to the new patient.

### 3.4 Closeness Among Patients

The closeness finding among patient's clusters them into three schemes within the data: Group1 clusters those highly adhered patients, who are characterized by being active and continuously following the assigned activity (see Algorithm2 and Figure2), Group 2 clusters different adherent patients, who are characterized by being active, passive or neutral (see Algorithm3 and Figure3). Finally, Group 3 measures closeness to people nearby. This is characterized by checking existing patient's data, then checking the new registered patient and calculating the distance between them. We use clustering in the first two methods and simple distance calculations in third scheme (see Algorithm4 and Figure4). When a new patient joins, the system measures distance with other patients to cluster them, then compare the new patient information to the highly adherent patients. Based on patient's lifestyle data, we create clusters and assign the relevant physical activity per cluster. The clusters and activity type the patients perform are stored with their means. For a new patient, we find the difference with the mean of the clusters and then closest clusters is chosen. To find the differences we use the lifestyle data and find nearest Euclidean distance from each mean of previous clusters.

---

**Algorithm 2: Cluster Based on Closeness to High Adherent Patients**

1  HighAdherent (P; C)
2      MaxIters(Limit of iteration);
3      $C = \{c_1; c_2; \ldots; c_k\}$ (set of cluster centroids);
4      $Q = \{PA_1; PA_2; \ldots; PA_m\}$ (list of physical activities);
5      foreach $p_i \in P$ do
6          Pick $C = \{c_1; c_2; \ldots; c_k\} \in$ High adherence people
7          ;
8          Algorithm 1;
9      end
10     return $PA_1$

---

With the first two algorithms we calculate differences with the mean of the clusters and the difference with the mean and the closest cluster is chosen. Since we know the clusters and activity type patients involved within a cluster and stored with patient's means, after finding new patient's clusters we list candidate physical activities from each cluster where the new patient was found and from the k nearby patient's the list is further narrowed by eliminating similar PAs. The final list of candidate PAs will be recommended to the coach to decide whether to assign them or not. The coach then enters the activity planning phase where he creates several schedules for the patient, help them set goals and other activities that could enhance their adherence.

## Algorithm 3: Cluster Based on Different Adherent Patients

```
1  DifferentAdherence (P; C)
      Input   :P=fp1 ; p2 ; : : : ; png (set of patient's to be clustered)
2     MaxIters(Limit of iteration);
3     C=fc1 ; c2 ; : : : ; ck g (set of cluster centroids)
4  c1 = cluster of high adherence group;
5  c2 = cluster of medium adherence group;
6  c3 = cluster of medium adherence group;
7  Q=fPA1 ; PA2 ; : : : ; PAmg (list of physical activities);
      Output  :PA2 2 Q (list of Physical activities in the cluster of different adherence group
8     foreach pi 2 P do
9         Pick C=fc1 ; c2 ; : : : ; ck g 2 different adherence people
10            ;
11        Algorithm 1;
12    end
13    return PA2
```

## Algorithm 4: Combined Physical Activity Recommendation

```
1  PARecommendation (P; C)
      Input   :P=fp1 ; p2 ; : : : ; png (set of patient's to be clustered)
2     K(number of clusters);
3     MaxIters(Limit of iteration);
4     C=fc1 ; c2 ; : : : ; ck g (set of cluster centroids);
5     Q=fPA1 ; PA2 ; : : : ; PAmg (list of physical activities in clusters of adherence groups);
6  PA=fPA1 ; PA2 ; : : : ; PAmg (list of physical activities);
7  PA  Q;
      Output  :l(PA) 2 Q (list of candidate Physical activities
8     foreach pi 2 P do
9         foreach pi 2 P do
10            Algorithm 2;
11        end
12        foreach pi 2 P do
13            Algorithm 3;
14        end
15        foreach pi 2 P do
16            KNN; (K nearest neighbours);
17        end
18        return pa 2 PA (list of Physical activities knns do);
19        l(PA)  f
20    end
21    while Q, f do
22        l(PA )  PA1 [ PA2 [ pa
23    end
```

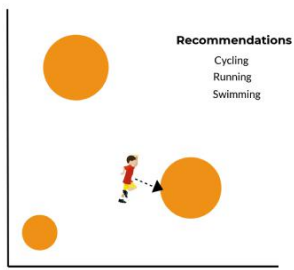

Fig. 2: Cluster Based on Closeness to High Adherent Patients

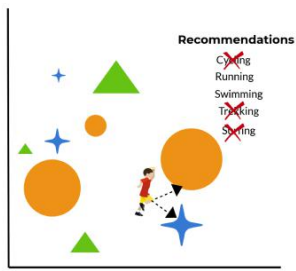

Fig. 3: Cluster Based on Different Adherent Patients

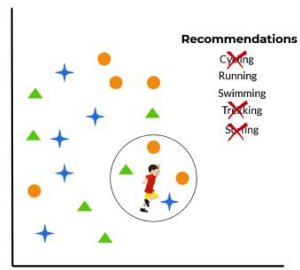

Fig. 4: Combined Physical Activity Recommendation

## 4  Conclusion

With rising costs of healthcare services, the current focus has shifted the preferences of individuals from health treatment to disease prevention. Existing telemedicine approaches in disease prevention all focus on handling patient's condition efficiently and recommend activities and plans for them in order to improve their health condition. We focused on increasing the efficiency of health services by first focusing on physician's workload and help decrease their fatigue during the diagnoses and patient follow up. Having most of the patient's spending time on their messaging apps makes using a chatbot application extremely easy. Applying conversational agents as a tool to facilitate physicians' tasks and provide follow up for patient can significantly improve patient engagement with the personal lifestyle coaching system and decrease physician's workload. We presented the activity recommender model algorithm that is formed by 1+3 chunks of algorithm to cluster patients, compare them with previous patient's and activities and hence match the recommendation to the coach. Our approach focuses on providing the recommendation to the coach rather than the patients. The approach will be validated with an ambulatory and the prototype will be tested with real users to measure the soundness of our approach.